# Layer-to-Layer Knowledge Mixing in Graph Neural Network for Chemical Property Prediction

Teng Jiek See,[a] Daokun Zhang,*[b] Mario Boley*[c,d] and David K. Chalmers*[a]

[a]Medicinal Chemistry, Monash Institute of Pharmaceutical Sciences, Monash University, 381 Royal Parade, Parkville, VIC 3068, Australia

[b]School of Computer Science, University of Nottingham Ningbo China, 199 Taikang East Road, Ningbo, 315100, China

[c]Department of Information Systems, University of Haifa, 65 Hanamal Street, Haifa 3303220, Israel

[d]Department of Data Science and AI, Faculty of Information Technology, Monash University, Clayton Campus, Building 63, 25 Exhibition Walk, VIC 3800, Australia

*Corresponding author

**Abstract**

Graph Neural Networks (GNNs) are the currently most effective methods for predicting molecular properties but there remains a need for more accurate models. GNN accuracy can be improved by increasing the model complexity but this also increases the computational cost and memory requirement during training and inference. In this study, we develop Layer-to-Layer Knowledge Mixing (LKM), a novel self-knowledge distillation method that increases the accuracy of state-of-the-art GNNs while adding negligible computational complexity during training and inference. By minimizing the mean absolute distance between pre-existing hidden embeddings of GNN layers, LKM efficiently aggregates multi-hop and multi-scale information, enabling improved representation of both local and global molecular features. We evaluated LKM using three diverse GNN architectures (DimeNet++, MXMNet, and PAMNet) using datasets of quantum chemical properties (QM9, MD17 and Chignolin). We found that the LKM method effectively reduces the mean absolute error of quantum chemical and biophysical property predictions by up to 9.8% (QM9), 45.3% (MD17 Energy), and 22.9% (Chignolin). This work demonstrates the potential of LKM to significantly improve the accuracy of GNNs for chemical property prediction without any substantial increase in training and inference cost.

## Introduction

Graph neural networks (GNNs) have emerged as the leading deep learning method for molecular property prediction{Wieder, 2020 #251}. GNNs encode molecular properties by representing molecules as sets of nodes and edges [1] using a message passing mechanism to update the feature vectors of each atom by considering electronic interactions with neighbouring atoms (Figure 1). The message passing mechanism enables GNNs to model complex interactions including the quantum mechanical (QM) interactions between atoms. Improved property predictions provided by GNNs can benefit diverse applications of chemistry. Recent applications of GNNs to chemical problems include predictions of: the solubility of pharmaceutical drugs[2], the potential for compounds to have mitochondrial toxicity[3], the carcinogenicity of small molecules[4], whether a molecule is able to permeate through the blood brain barrier[5] and to construct efficient and accurate molecular mechanics forcefields[6].

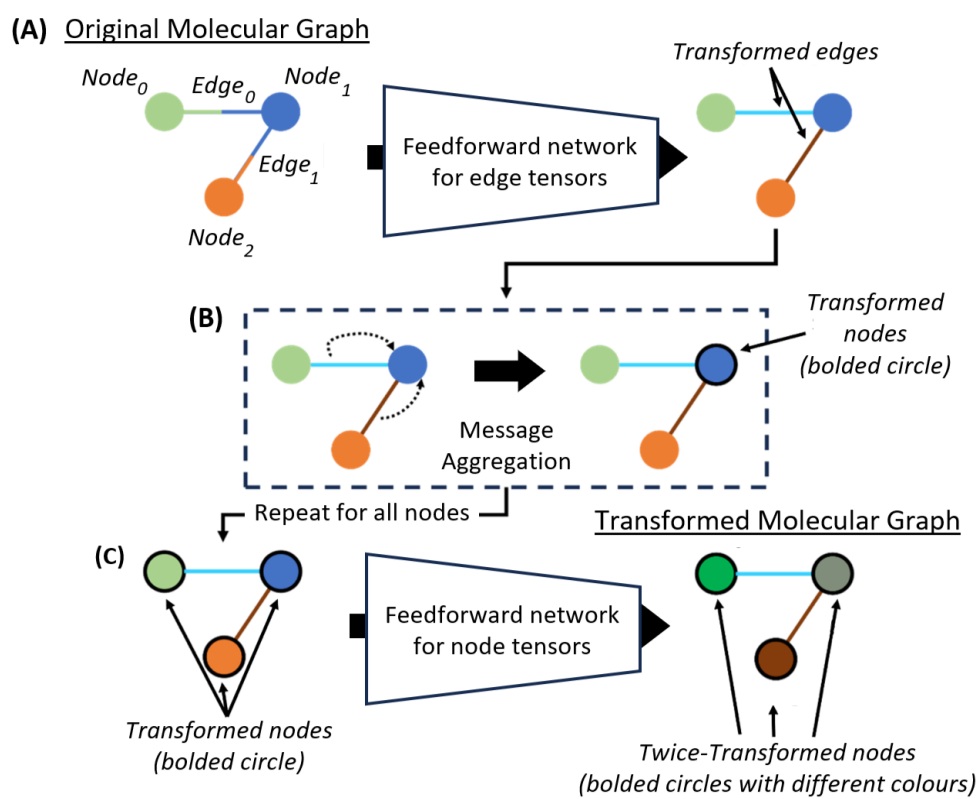

*Figure 1. A Graph Neural Network (GNN) layer is a fundamental building block of GNNs. GNN layers use message passing mechanism to process molecular information. (A) A molecular graph is input into a feedforward network to encode the connections between atoms in the form of edge feature tensors. (B) Each node aggregates edge feature tensors from its neighbouring nodes via a process called message aggregation. (C) The node feature tensors are updated using another feedforward network to reflect the newly aggregated information.*

One important application of GNNs is prediction of small molecule quantum chemical properties. For example, molecular energies, geometries or polarizability. Until the past few years, such properties could only be calculated using quantum mechanics (QM) calculations of the underlying electronic structure of the molecular system of interest. A large number of methods can be used to calculate these properties, (e.g. density functional theory (DFT) calculations); however, these calculations are all complex and expensive. More recently, it has been shown that neural networks trained on QM calculations effectively predict molecular properties. For instance, a family of artificial neural networks (e.g. ANI-1x{Smith, 2018 #10} and ANI-2x{Devereux, 2020 #8}) and GNNs (e.g. DimeNet[7] and SphereNet[8]) have been trained on large datasets of QM calculations to predict molecular properties. These models haven been used in virtual screening protocol to improve binding pose scoring[9] and simulation of inorganic materials[10, 11]. Although GNN prediction of chemical properties is advancing, there remains significant room for improvement in prediction accuracy. For instance, accurate prediction of reaction energies during pathway exploration using Artificial Force Induced Reaction (AFIR) in complex systems remains unattainable without extensive fine-tuning of GNNs on specialised datasets [12]. Similarly, GNN-based molecular dynamics simulations frequently lose stability within just a few hundred picoseconds [13] and reliable nuclear magnetic resonance (NMR) predictions require retraining GNN on large datasets [14]. Collectively, these issues highlight a critical gap in the ability of GNNs to fully understand the geometry and topology of molecular systems. Overcoming these limitations is crucial for GNNs to realise its full potential in chemistry. Recent advances in GNN accuracy have largely resulted from bigger and more complex models, with increasing hidden embedding dimensions and more complex message passing mechanisms, increasing the complexity and computational cost of the predictions[15, 16, 17]. Therefore, there is a need for improved GNN methods that improve model accuracy while not increasing computational cost. Such improvements will be important to enable large-scale molecular simulations and property predictions using GNN models.

In deep learning, knowledge distillation[18] is a method to improve neural network computational efficiency by transferring knowledge from a large teacher model to a smaller student model. Knowledge distillation creates a lightweight model that performs closely to the larger teacher model but requiring lower computational power and memory at prediction time. A notable application of knowledge distillation is Multi-Scale Knowledge Distillation (MSKD) [19]. MSKD utilizes multiple large teacher GNN models and transfers topological information to a single student GNN model. The teacher models are designed to have varying numbers of hidden layers ($l$-layers), which capture knowledge at different scales of topological semantics. By distilling information from multiple teachers, the student GNN model achieves accuracy beyond what it would attain through independent training alone. A limitation of MSKD is that training using multiple teacher GNN models incurs significant additional computational cost.

Self-knowledge distillation[20,21,22] is a specialized branch of knowledge distillation where a model increases its prediction accuracy by using its own outputs to refine its intermediate or final predictions. For example, Mixture-of-Distilled-Experts (MoDE)[23] is a mixture of experts model that uses mutual distillation among experts to distribute the knowledge learned by other experts. MoDE enhances the understanding of the individual experts and overall boosts the model's ability to generalize. Unlike standard knowledge distillation, self-knowledge distillation does not rely on external models.

In this work, we aimed to investigate the potential of using self-knowledge distillation to improve the prediction accuracy of GNNs in the field of chemical property prediction. Recognising that GNNs consist of multiple layers, we proposed that prediction accuracy could be improved by treating each GNN layer as an expert that captures knowledge at different scales, which could benefit information sharing by knowledge distillation between layers. This can be described as a form of self-knowledge distillation, where the same GNN model serves simultaneously as both teacher and student. Each layer acts as a teacher for another layer while also learning as a student from a different layer. Taking this proposition, we have designed a self-knowledge distillation method, which we call Layer-to-Layer Knowledge Mixing (LKM), which facilitates distillation between all pairs of layer embeddings. LKM treats each GNN layer as an expert with unique topological insights and promotes knowledge exchange between these layers. We investigate the ability of LKM for prediction accuracy improvement across a set of quantum chemical thermodynamic properties for small molecules and a larger biomolecular system. To determine the optimal distillation configuration, we test different distillation strengths and examine their effect on accuracy improvement. We also investigate how the number of teacher-student interactions in LKM, modulated by increasing the number of GNN layers, relates to improvements in prediction accuracy compared to GNNs without using LKM.

**Methods**

*Layer-to-Layer Knowledge Mixing*

State-of-the-art GNN models [24][25][7][26][8][27][28] use Deep Supervision[29] where each GNN layer contributes directly to the final output via a node processing block. Each layer node processing block combines information from its respective edge block to produce different hop information to improve overall accuracy. For an *l*-layer GNN, a neighborhood aggregation scheme is performed to capture the *l*-hop information surrounding each node, which makes the GNN model encode the topological semantic information at the scale of *l*-hop.[30][31][32] Since each GNN layer node processing block captures different hop information (e.g. $layer_1$ – 1 hop, $layer_2$ – 2 hops, etc.), it is possible to treat each layer node processing block as an expert or a separate sub-model that contains knowledge at different scale (Figure 2). The entire GNN model can therefore be treated as a Mixture of Experts (MoE).

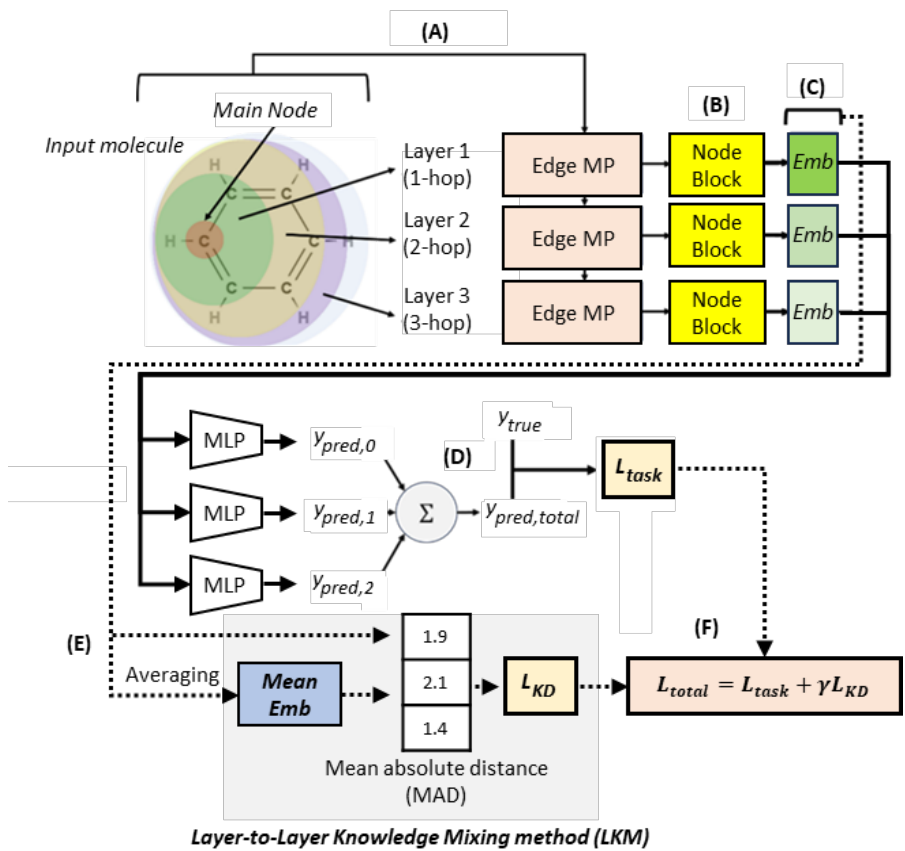

*Figure 2. The Layer-to-Layer Knowledge Mixing method incorporated into a generic Graph Neural Network framework. Benzene ($C_6H_6$) serves as an example input. (A) Benzene is fed into the three edge-based message passing (Edge MP) layers, each processes different hop radius. (B) Edges are aggregated into atomic nodes and node blocks process these nodes into embeddings. (C) Node embeddings are fed in a multi-layer perception (MLP) for regression tasks. (D) Predictions are compared to true values to produce prediction error term $L_{task}$. (E) LKM computes knowledge distillation loss $L_{KD}$ using the mean squared distance between node embeddings and their mean embedding. (F) Both $L_{task}$ and $L_{KD}$ (weighted via knowledge distillation strength $\gamma$) are summed together to create the final objective term $L_{total}$ for minimization.*

Inspired by MoDE, MSKD and Deep Supervision, we propose a self-knowledge distillation method called Layer-to-Layer Knowledge Mixing (LKM) that aims to boost the prediction accuracy of GNN models without significant computational cost. The LKM method uses the multi-scale property and deep-supervision architecture that is already present in current GNN models. Much like MoDE, we take the embedding $e_{i,j} \in \mathbb{R}^d$ from the node processing block for atom $i$ at layer $j$, and compute the average embedding $\bar{e}_i \in \mathbb{R}^d$ for atom $i$ over all layers, where $m$ is the number of GNN layers (Equation 1).

$$\bar{e}_i = \frac{1}{m}\sum_{i=1}^{m} e_{i,j} \qquad (Equation\ 1)$$

We then calculate the mean squared distance between each layer node processing block embedding $e_{i,j}$ and the average embedding $\bar{e}_i$ over all atoms, where $n$ is the number of atoms (Equation 2). Next, we take the average of all mean squared distances to obtain an embedding discrepancy value called the knowledge distillation loss term $L_{kd}$.

$$L_{kd} = \frac{1}{mn} \sum_{i=1}^{m} \sum_{j=1}^{n} \left\| \bar{e}_i - e_{i,j} \right\|_2^2 \qquad (Equation\ 2)$$

Finally, we add $L_{kd} \in \mathbb{R}^1$ into the total loss function $L_{total}$ as an auxiliary loss function to the main regression loss function (in this case, we use mean absolute error). $L_{kd}$ acts as a representation regularisation to the training objective that measure the embedding discrepancies between the layer embeddings and the average embedding. We add a weighting hyperparameter $\gamma$ (also known as the knowledge distillation strength) to vary the contribution of $L_{kd}$ (Equation 3)

$$L_{total} = \left( \frac{1}{n} \sum_{i=1}^{n} |y_i - \hat{y}_i| \right) + \gamma L_{kd} \qquad (Equation\ 3)$$

where $y_i$ is the truth property value of the *i-th* conformer, $\hat{y}_i$ is the corresponding prediction from the model and $n$ is the number of samples.

*Datasets*

The LKM method was evaluated on its ability to improve the prediction accuracy of molecular properties. We used the QM9[33], MD17[34] and Chignolin[35] datasets. The QM9 dataset consists of 12 molecular properties that have been calculated using high-quality quantum chemical methods. The properties are: dipole moment ($Mu$), isotropic polarizability (Alpha), highest occupied molecular orbital energy (HOMO), lowest unoccupied molecular orbital energy (LUMO), energy gap between HOMO and LUMO (Gap), electronic spatial extent ($<R^2>$), zero-point vibrational energy (ZPVE), internal energy at 0 K ($U_0$), internal energy at 298.15 K (U), enthalpy at 298.15 K (H), Gibbs free energy at 298.15 K (G) and heat capacity at 298.15 K ($C_v$). The values have been calculated for approximately 134,000 stable small organic molecules (Table 1). Each molecule is composed of up to nine heavy atoms (C, N, O, and F). The QM9 dataset was divided into three parts: a training set containing 110,000 molecules, a validation set with 10,000 molecules, and a test set comprising the remaining molecules.

The MD17 dataset consists of quantum chemical energy and forces calculated from the molecular dynamics trajectories of eight small organic molecules. The dataset consists of 15,770 to 875,237 conformers of each molecule (Table 1). This dataset was split into three sections: a training set composed of 1000 conformers, a validation set of 1000 conformers, and a test set containing remaining conformers.

The Chignolin data describes the macromolecule Chignolin (PDB ID: 5AWL), which has 166 atoms. It consists of 9543 conformations sampled by replica exchange MD[36] with a Berendsen thermostat[37] at 340K with energies calculated at M06-2X/6-31G* DFT level[38] using Gaussian 16 [39] (Table 1). The Chignolin dataset was split into three subsets: a training set (80%), a validation set (10%) and a test set (10%).

*Table 1. Datasets used to evaluate the LKM method.*

| Dataset | Number of molecular conformations/samples | | |
|---|---|---|---|
|  | Training | Validation | Test |
| QM9 | 110000 | 10000 | 23885 |
| MD17 | 1000 | 1000 | 507983 (benzene)<br>13770 (uracil)<br>206250 (naphthalene)<br>91762 (aspirin)<br>200231 (salicylic acid)<br>873237 (malonaldehyde)<br>435092 (ethanol)<br>442790 (toluene) |
| Chignolin | 7634 | 954 | 955 |

*Model training*

We evaluated the LKM method using three GNN models: DimeNet++[8], MXMNet[8] and PAMNet[8] which were obtained from Liu et al.[8], Zhang et al.[27] and Zhang et al.[28] respectively. The LKM method was implemented using PyTorch.[40] Models were trained by minimizing the mean absolute error loss using the Adam optimizer [41] with the specific parameters: $learning\ rate = 5 \times 10^{-4}$, $\beta_1 = 0.9$, $\beta_2 = 0.999$, and $\varepsilon = 10^{-8}$. Additionally, a linear learning rate warm-up[42] was used for either 3000 (QM9 and Chignolin) or 1000 steps (MD17), followed by a systematic reduction in the learning rate through a cosine annealing decay. The training processes spanned 150 epochs with a batch of 32 for QM9, for 2000 epochs with a batch size of 8 for MD17 and for 100 epochs with a batch size of 2 for Chignolin. We incorporated a hybrid energy-force error loss function using the hyperparameters described by Klicpera et al.[26] to train the GNN models on both energy and force labels on MD17.

GNN model performance was evaluated using two metrics. First, we used mean absolute error (MAE or $E_{mae}$),

$$E_{mae} = \frac{1}{n}\sum_{i=1}^{n}|y_i - \hat{y}_i| \qquad \text{(Equation 4)}$$

where $y_i$ is the truth property value of the *i-th* molecule/conformer, $\hat{y}_i$ is the corresponding prediction from the model and *n* is the number of samples.

Second, we calculated the percentage reduction in mean absolute error when using LKM compared to not using it ($\Delta E_{mae}$),

$$\Delta E_{mae} = \frac{E_{mae,LKM} - E_{mae,base}}{E_{mae,base}} \times 100\% \qquad \text{(Equation 5)}$$

where $E_{mae,LKM}$ is the MAE of a model that uses our LKM method and $E_{mae,base}$ is the MAE of a model without using our LKM method.

# Results

*Applicability of Layer-to-Layer Knowledge Mixing across GNN models*

Our first benchmark evaluated the LKM method on small molecule quantum property prediction using three GNN models, DimeNet++[8], MXMNet[8] and PAMNet[8]. These three GNN models represent multiple generations of GNNs architectures that employ different mechanisms for molecular property prediction. Using a diverse set of GNN architectures is a robust way to evaluate the ability of the LKM method in improving prediction accuracy regardless of the underlying architectural differences among the GNNs. All three GNN models were trained under the same configuration using the QM9 dataset. The LKM method achieved large improvements in almost all 12 QM9 prediction tasks (Figure 3). Specifically, it reduced the overall average MAE of PAMNet, MXMNet and DimeNet++ by 15.2%, 9.8%, and 1.3% respectively. The average improvement in $\Delta E_{mae}$ was notably larger for PAMNet and MXMNet than for DimeNet++.

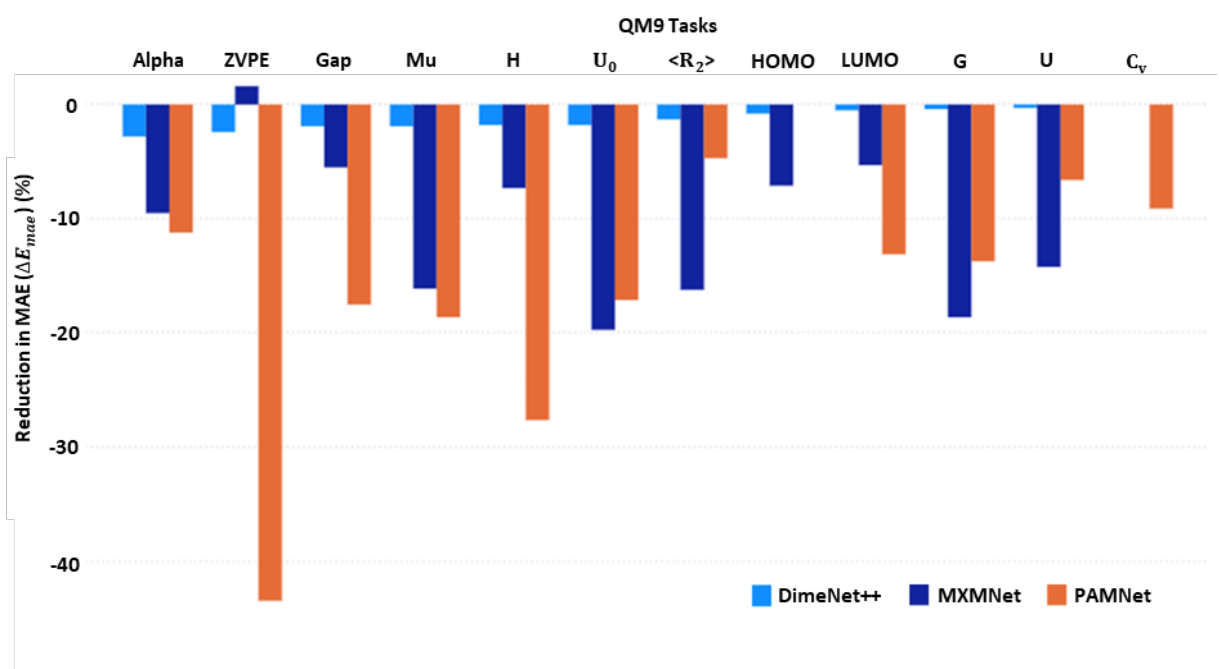

*Figure 3. Reduction in prediction error for the QM9 dataset using Layer-to-Layer Knowledge Mixing (LKM) Method in GNN models (DimeNet++, MXMNet, and PAMNet). LKM method significantly enhances prediction accuracy compared to the baseline in nearly all tasks, particularly Alpha, Gap, Mu, and H. Detailed prediction error data is provided in the Supporting Information (Table S1).*

*Molecular dynamics trajectory energies and forces of small molecules*

Molecular dynamics simulations involve calculating the movements of atoms and molecules by determining their interaction energies and the corresponding forces acting on them. Molecular interactions can be predicted using GNNs. We evaluated the ability of our LKM method to improve GNN prediction of molecular energies and forces from molecular dynamics trajectory data using the MD17 dataset[34]. This dataset consists of structures sampled from molecular dynamics trajectories of 8 small molecules. The LKM method was evaluated within DimeNet++ and MXMNet. For each molecule, we used a training set comprising 1000 conformations and a validation set of equal size (Figure 4). Our LKM method substantially improved DimeNet++ predictions of molecular energies and forces, giving average MAE reductions of 26.20% in energy and 13.69% in force prediction. Similarly, applying LKM to MXMNet improved 11 out of 12 tasks in energy prediction and 10 out of 12 in force prediction. The overall average $\Delta E_{mae}$ for MXMNet with LKM was -45.29% in energy and -2.96% in force prediction.

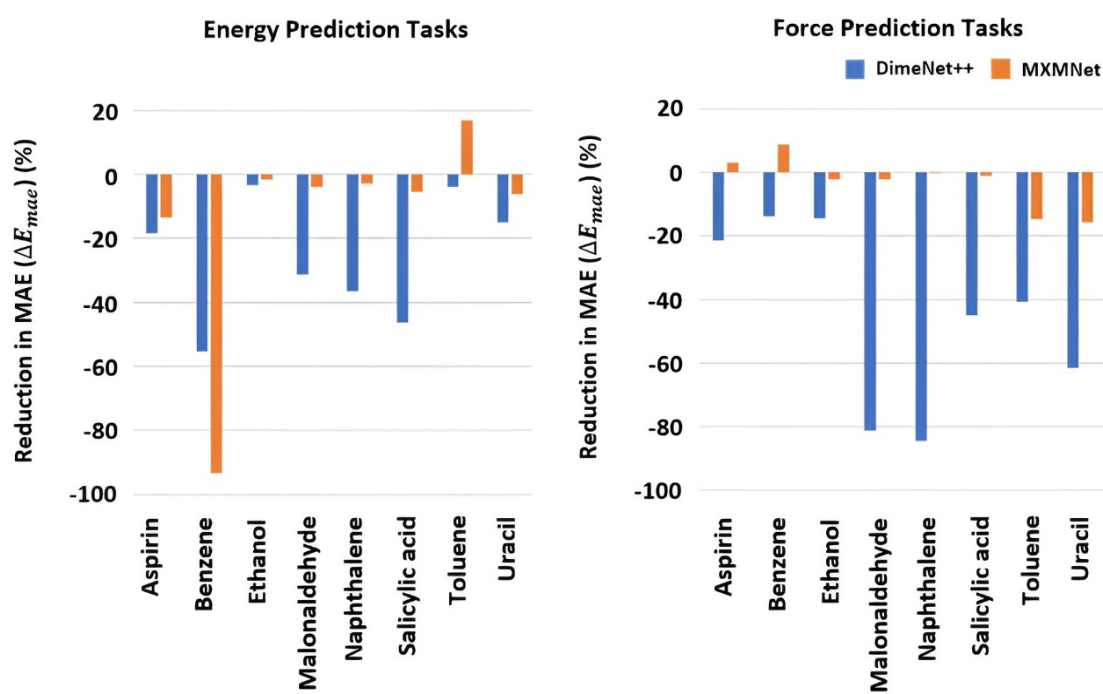

*Figure 4. Evaluation of molecular dynamics trajectory energies and forces prediction using the MD17 dataset. Our layer-to-layer knowledge mixing method (LKM) is assessed using DimeNet++ and MXMNet. When integrated with DimeNet++, LKM consistently reduced $E_{mae}$ for all 12 molecules in both energy and force prediction categories, with significant improvements observed for Benzene, Malonaldehyde, and Naphthalene. Similarly, when applied to MXMNet, LKM reduced $E_{mae}$ for the majority of the molecules in both categories, with notable gains for Benzene and Uracil. Detailed prediction error data can be found in the Supporting Information (Table S2).*

*Protein conformational energy prediction*

We evaluated the ability of our LKM aggregation method to boost energy prediction accuracy of larger molecular systems using the Chignolin dataset[40]. This dataset consists of conformations sampled from molecular dynamics trajectories of the 10 amino acid peptide chignolin. We evaluated LKM with MXMNet and PAMNet using a training set comprising 80% of the molecules, a validation set of 10%, and a test set of 10%. Our results (Table 2) show that the LKM method reduced MAE in predicting relative conformational energies for both MXMNet (-2.2%) and PAMNet (-22.9%). Additionally, we applied molecular mechanics using the OPLS4 force field [43], as implemented in Schrodinger Maestro [44], across the test conformations. OPLS4 achieved the worst accuracy compared to the GNN models (MXMNet and PAMNet) which is expected because OPLS4 is not trained on the Chignolin dataset. Overall, our LKM method greatly reduced the MAE energy prediction for the peptide system.

*Table 2. Protein energy conformational prediction error for the Chignolin Protein dataset. Our LKM method is evaluated using MXMNet and PAMNet. Our LKM method (knowledge distillation strength $\gamma = 0.002$) improved energy calculation for both models and is more substantially accurate for PAMNet. Both GNN models surpassed OPLS4 in accuracy. Values in square brackets [ ] are the percentage decrease in MAE given by LKM method over the base model.*

| Mean absolute error (eV) (Lower is better) | | | | |
|---|---|---|---|---|
| Molecular mechanics (OPLS4) | MXMNet | | PAMNet | |
| | Base | Base + LKM (ours) | Base | Base + LKM (ours) |
| 4.17 | 0.58 | 0.57 [-2.2%] | 0.99 | 0.77 [-22.9%] |

*Knowledge distillation strength and its effect on prediction accuracy*

The LKM method incorporates a knowledge distillation strength term ($\gamma$) that controls the trade-off between the prediction loss function and the knowledge distillation loss function during training. We investigated how $\gamma$ affected prediction accuracy using MXMNet using the QM9 dataset. The $\gamma$ term was varied between 0.0 and 100, while the learning rate schedules and batch sizes were kept constant. The average reduction in MAE $\Delta E_{mae}$ of MXMNet decreased, reaching a minimum of -7.89% when $\gamma$ = 0.02 and then increased onwards (see Figure 5). A high $\gamma$ around 1 resulted in a 1.78% average $\Delta E_{mae}$ while a higher $\gamma$ around 100 led to a 15.66% increase in average $\Delta E_{mae}$. Therefore, the optimal $\gamma$ value is 0.02.

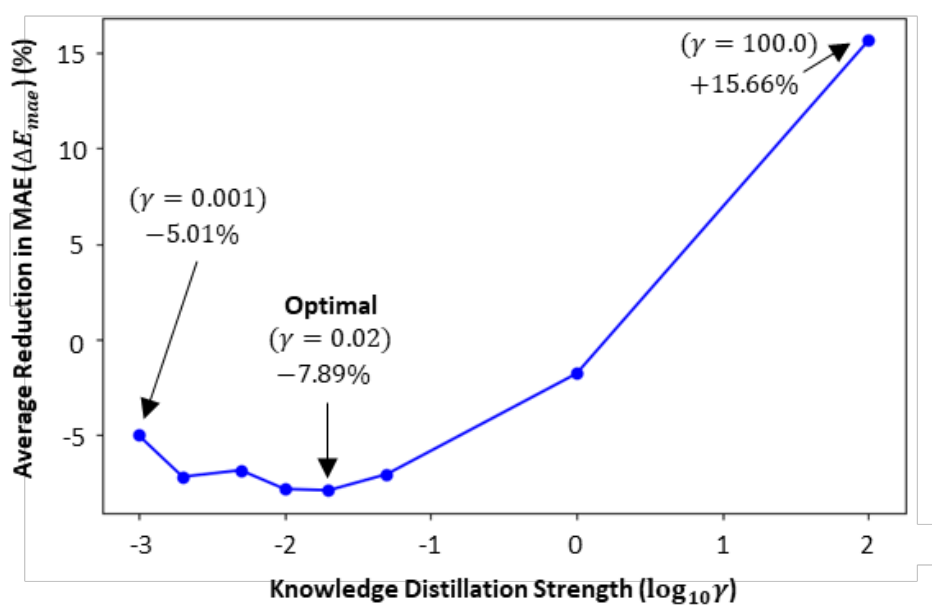

*Figure 5. Evaluation of the Layer-to-Layer Knowledge Mixing Method (LKM) across various Knowledge Distillation Strengths ($\gamma$) Using MXMNet on the QM9 Dataset. The average $\Delta E_{mae}$ decreased gradually from -5.01% ($\gamma$ = 0.001), peaks at -7.89% ($\gamma$ = 0.02), and then increased beyond that point. Detailed prediction error data can be found in the Supporting Information (Table S3).*

*Effects of LKM method on varying number of GNN layers*

The number of layers in a GNN model correlates with its ability to learn complex atomic interactions in chemical graph structures. We investigated how our LKM method affects the prediction accuracy of the GNN model across different numbers of layers. Using MXMNet as the base model, we trained it on 12 thermodynamic property tasks of the QM9 dataset with 2, 4, and 6 layers separately. All models had knowledge distillation strength $\gamma = 0.05$. Our LKM method reduced the average MAE across all layer configurations, with the improvement ranging from -5.71% to -7.11% (Figure 6). Interestingly, the average MAE reduction increased as the number of layers increased. Overall, our LKM method enhances the accuracy of the GNN model regardless of the number of layers used, with the best accuracy observed with the largest number of layers.

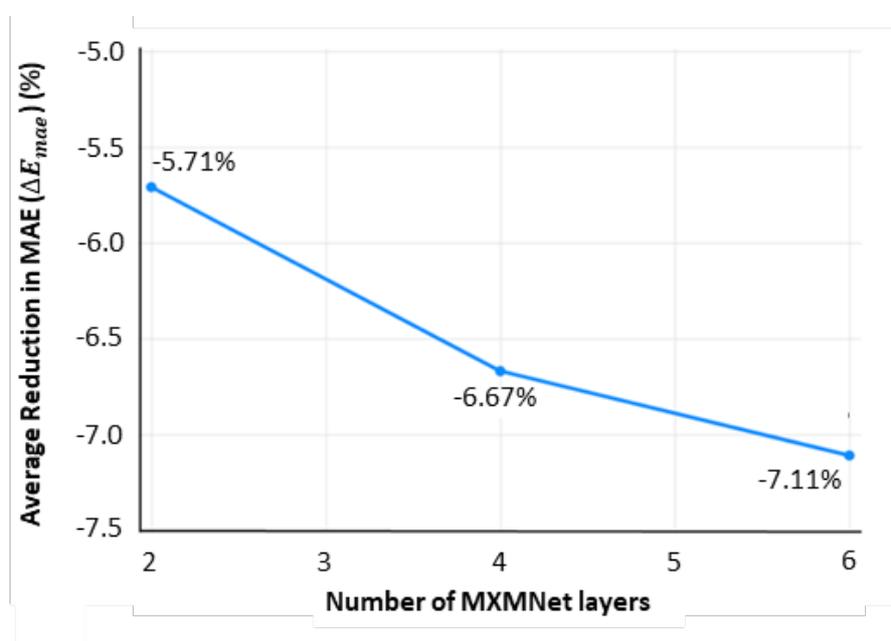

*Figure 6. Average reduction of prediction error across different numbers of GNN layers in MXMNet with the layer-to-layer knowledge mixing (LKM) method is applied at a fixed knowledge distillation strength ($\gamma = 0.05$). This comparison utilizes the QM9 dataset. The LKM method consistently reduces MAE regardless of the number of layers. Detailed data on prediction errors can be found in the Supporting Information (Table S4).*

**Discussion**

In this work, we have developed Layer-to-Layer Knowledge Mixing (LKM), a self-knowledge distillation method that enhances the accuracy of Graph Neural Networks (GNNs) and show that it substantially improves predictions of quantum molecular properties. The LKM method was inspired by Mixture-of-Distilled-Experts (MoDE) and Multi-Scale Knowledge Distillation (MSKD). State-of-the-art GNN models have layers can be interpreted as individual experts in a mixture of experts framework. Each layer processes and learns information at different hop radii. By facilitating knowledge distillation across these layers, LKM enables the transfer of information between different hop radii without the need for additional teacher models and extra parameters.

Importantly, the LKM method does not introduce any new parameters to the GNN model and therefore does not incur major computational cost. This makes the LKM method a computationally efficient approach, significantly improving GNN accuracy, without increasing the model complexity or runtime. We have also shown that the LKM method can be readily incorporated into existing GNNs to improve their performance.

We evaluated the LKM method on three state-of-the-art GNN models (DimeNet++, MXMNet and PAMNet), testing molecular property prediction using the QM9, MD17 and Chignolin quantum chemical datasets. We found that the LKM method consistently improved the ability of GNNs to predict a variety of chemical properties (molecular energy, atomic forces and thermodynamic properties) across molecular sizes ranging from 3 atoms to 166 atoms in size. In the domain of small molecules, our LKM method reduced the average $E_{mae}$ of DimeNet++ (-1.3%), MXMNet (-9.8%) and PAMNet (-15.2%) in predicting quantum chemical thermodynamic properties of QM9 across the majority of the property prediction tasks in total (Table S1 and Figure 3). Similarly, our LKM method also improved prediction of small molecule energies and forces. Using the MD17 dataset of 8 different organic molecules (Table S2 and Figure 4), we found that LKM reduced the $E_{mae}$ of DimeNet++ (-26.2% for energy and -13.7% for force) and MXMNet (-45.3% for energy and -29% for force). In the realm of biomolecules, the LKM method reduced the $E_{mae}$ of MXMNet (-2.2%) and PAMNet (-22.9%) in predicting the conformational energies of Chignolin (Table 1). These consistent improvements in prediction accuracy show that distilling different multi-hop/multi-scale information between each layer helps in learning more complex patterns and leads to better generalisation.

The knowledge distillation strength term ($\gamma$) determines the balance between the auxiliary knowledge distillation loss function and the main prediction task loss function during training. Adjusting $\gamma$ influences prediction accuracy. For instance, increasing $\gamma$ from 0 to 0.02 results in $\Delta E_{mae}$ of -7.89%. However, values of $\gamma$ beyond 0.02 lead to an increase in $\Delta E_{mae}$, increasing by +15.66% when $\gamma$ reaches 100. This increased in prediction error occurs because larger values of $\gamma$ cause all layers to converge to the same representation, which undermines the multi-view/multi-hop information deep supervision prediction power (Figure 5 and Table S3). A similar effect was found with MoDE[23] where increasing the knowledge distillation strength term initially boosted accuracy but beyond a certain point, further increase caused the accuracy to plateau and eventually decline. Therefore, identifying the optimal $\gamma$ value is important to obtain the best accuracy.

We also investigated how the number of layers in the GNN affected the LKM method. We tested 2, 4, and 6 layers separately and found in all cases that LKM improved prediction accuracy. Further that accuracy improvement increased with the number of layers (Figure 6 and Table S4). This observation is consistent with findings of Zhang et al [19], Wang et al [45] and Zaras et al [46] that multi-teacher approaches can boost student prediction accuracy. Using the LKM method, each GNN layer effectively acts as a teacher that imparts its own learned hop radius information about the molecule to the student, resembling a multi-teacher and one-student setup. Since each GNN layer is both a teacher and a student, increasing the number of GNN layers increases the number of teachers and students which explains why the accuracy improves when the number of GNN layers increases. More importantly, unlike the mentioned knowledge distillation approaches [19,45,46], our LKM approach does not add additional parameters or computation because it capitalises on the varying perspectives each layer learns during training.

**Conclusion**

In this work, we aimed to improve the performance of Graph Neural Network (GNN) models in predicting quantum chemical properties (e.g. molecular energy) without increasing major computational cost or complexity of the model. Achieving this, we have developed Layer-to-Layer Knowledge Mixing (LKM), a novel self-knowledge distillation method, that increases the accuracy of state-of-the-art quantum chemistry GNNs by distilling multi-hop/multi-scale information across different layers of GNNs, without the use of additional teacher models.

We evaluated LKM on molecular datasets ranging from 3 to 166 atoms in size with properties calculated by standard quantum chemical (DFT) methods. In each case, we found that LKM notably predications of diverse quantum chemical properties. Specifically, considering all 12 properties of the QM9 dataset, incorporation of LKM into PAMNet, MXMNet and DimeNet++ gave average MAE reductions of -15.2%, -9.8% and -1.3%, respectively. We also showed that the knowledge distillation strength parameter ($\gamma$) and the number of GNN layers are important for optimizing model accuracy. The $\gamma$ value is sensitive with the GNN prediction accuracy such that its accuracy is the highest within an optimal range and degrades when the $\gamma$ value is set too high or too low. Increasing the number of GNN layers generally improve accuracy.

Unlike other knowledge distillation methods for graph neural networks such as Mulde [45] and MSKD [19], our LKM method enhances prediction accuracy without adding significant computational cost. By sharing the existing varying perspectives learned by each layer, LKM provides a zero-cost method for improving GNN accuracy, which can easily be applied to improve the accuracy of molecular property predictions. The LKM method also has potential for further improvement. One approach might be to incorporate an adaptive mechanism to selectively distil knowledge from teacher to student GNN layer rather than accepting all teacher information uniformly. More targeted information transfer could potentially further improve the effectiveness of our method.

**Supporting Information**

The code that we used and developed can be found in [tengjieksee/Layer-to-Layer-Knowledge-Mixing-Graph-Neural-Network-Official: The official repository of Layer-to-Layer-Knowledge-Mixing-Graph-Neural-Network-Official.](#)

**Supporting Information**

*Table S1. Evaluation of layer-to-layer knowledge mixing method (LKM). Three GNN models were evaluated: DimeNet++, MXMNet and PAMNet. Comparison uses the QM9 dataset. Our LKM method improves the mean prediction accuracy over baseline in the majority of tasks. Values in square brackets [ ] are the MAE reduction $\Delta E_{mae}$ given by LKM method over the base model.*

| | | Mean absolute error (lower is better) | | | | | |
|---|---|---|---|---|---|---|---|
| | GNN model | DimeNet++ | | MXMNet | | PAMNet | |
| Task | Unit | Base | Base+LKM (ours) | Base | Base+LKM (ours) | Base | Base+LKM (ours) |
| Mu | D | 0.0267 | **0.0262** [-1.9%][a] | 0.0384 | **0.0322** [-16.1%][a] | 0.0429 | **0.0349** [-18.6%][c] |
| Alpha | $a_0^3$ | 0.0424 | **0.0412** [-2.8%][a] | 0.0525 | **0.0475** [-9.5%][a] | 0.0724 | **0.0643** [-11.2%][b] |
| HOMO | meV | 23.9 | **23.7** [-0.8%][a] | 23.8 | **22.1** [-7.1%][a] | **32.9** | 32.9 [+0.0%][c] |
| LUMO | meV | 18.3 | **18.2** [-0.5%][a] | 18.9 | **17.9** [-5.3%][a] | 31.2 | **27.1** [-13.1%][a] |
| Gap | meV | 41.3 | **40.5** [-1.9%][c] | 40.1 | **37.9** [-5.5%][a] | 60.4 | **49.8** [-17.5%][a] |
| $<R^2>$ | $a_0^2$ | 0.317 | **0.313** [-1.3%][a] | 0.530 | **0.444** [-16.2%][a] | 0.772 | **0.736** [-4.7%][a] |
| ZPVE | meV | 1.24 | **1.21** [-2.4%][a] | **1.24** | 1.26 [+1.6%][a] | 3.34 | **1.89** [-43.4%][a] |
| $U_0$ | meV | 6.22 | **6.11** [-1.8%][a] | 8.99 | **7.22** [-19.7%][a] | 12.56 | **10.41** [-17.1%][a] |
| U | meV | 6.11 | **6.09** [-0.3%][b] | 8.53 | **7.32** [-14.2%][a] | 13.74 | **12.84** [-6.6%][a] |
| H | meV | 6.14 | **6.03** [-1.8%][c] | 7.78 | **7.21** [-7.3%][a] | 11.36 | **8.22** [-27.6%][c] |
| G | meV | 6.98 | **6.95** [-0.4%][a] | 9.73 | **7.92** [-18.6%][a] | 11.81 | **10.19** [-13.7%][a] |
| $C_v$ | $\frac{cal}{mol\ K}$ | **0.023** | 0.023 [+0.0%][c] | **0.024** | 0.024 [+0.0%][a] | 0.033 | **0.030** [-9.1%][a] |
| **Number of winning tasks** | | 0.5/12 | 11.5/12 | 1.5/12 | 10.5/12 | 0.5/12 | 11.5/12 |
| **Average $\Delta E_{mae}$ (%)** | | - | -1.34% | - | -9.83% | - | -15.23% |

[a]$\gamma = 0.002$. [b]$\gamma = 0.005$. [c]$\gamma = 0.01$

*Table S2. Calculation errors for molecular energies for the MD17 dataset. Our layer-to-layer knowledge mixing method (LKM) is evaluated using DimeNet++ and MXMNet. DimeNet++ with LKM won all 12 tasks in both Energy and Force prediction categories, with average MAE reduction of 26.20% and 13.69%, respectively. MXMNet with LKM won 11 out of 12 Energy tasks and 9.5 out of 12 Force prediction tasks, with average accuracy improvements of 45.29% and 2.96%, respectively. Values in square brackets [ ] are the average $\Delta E_{mae}$ given by LKM method over the base model.*

| | | Mean absolute error (lower is better) | | | |
|---|---|---|---|---|---|
| | GNN models | DimeNet++ | | MXMNet | |
| Molecules | Tasks | Base | Base+LKM (ours) | Base | Base+LKM (ours) |
| Aspirin | Energy (kcal/mol) | 0.211 | **0.173 [-18.2]**[b] | 0.616 | **0.534 [-13.3%]**[b] |
| | Force (kcal/mol/Å) | 0.477 | **0.375 [-21.4%]**[b] | **0.689** | 0.711 [+3.1%][b] |
| Benzene | Energy (kcal/mol) | 0.358 | **0.160 [-55.3%]**[b] | 1.026 | **0.066 [-93.5%]**[d] |
| | Force (kcal/mol/Å) | 0.206 | **0.178 [-13.7%]**[b] | **0.186** | 0.202 [+8.8%][d] |
| Ethanol | Energy (kcal/mol) | 0.058 | **0.056 [-3.3%]**[b] | 0.069 | **0.068 [-1.4%]**[b] |
| | Force (kcal/mol/Å) | 0.174 | **0.149 [-14.5%]**[b] | 0.271 | **0.266 [-2.0%]**[b] |
| Malonaldehyde | Energy (kcal/mol) | 0.128 | **0.0879 [-31.3%]**[c] | 0.106 | **0.102 [-3.8%]**[b] |
| | Force (kcal/mol/Å) | 1.418 | **0.266 [-81.2%]**[c] | 0.376 | **0.368 [-2.1%]**[b] |
| Naphthalene | Energy (kcal/mol) | 0.184 | **0.117 [-36.5%]**[b] | 0.894 | **0.868 [-2.9%]**[d] |
| | Force (kcal/mol/Å) | 0.806 | **0.125 [-84.5%]**[b] | 0.231 | **0.230 [-0.1%]**[d] |
| Salicylic acid | Energy (kcal/mol) | 0.299 | **0.161 [-46.2%]**[c] | 3.769 | **3.567 [-5.4%]**[b] |
| | Force (kcal/mol/Å) | 1.101 | **0.607 [-44.9%]**[c] | 0.480 | **0.475 [-1.1%]**[b] |
| Toluene | Energy (kcal/mol) | 0.099 | **0.095 [-3.9%]**[b] | **0.794** | 0.929 [+16.9%][d] |
| | Force (kcal/mol/Å) | 0.182 | **0.108 [-40.8%]**[b] | 0.246 | **0.210 [-14.6%]**[d] |
| Uracil | Energy (kcal/mol) | 0.129 | **0.109 [-14.9%]**[b] | 0.113 | **0.106 [-6.2%]**[a] |
| | Force (kcal/mol/Å) | 0.482 | **0.186 [-61.4%]**[b] | 0.232 | **0.196 [-15.6%]**[a] |
| Number of winning tasks | Energy | 0/12 | **12/12** | 1/12 | **11/12** |
| | Force | 0/12 | **12/12** | 2/12 | **10/12** |
| Average $\Delta E_{mae}$ (%) | Energy | - | **-26.20%** | - | **-45.29%** |
| | Force | - | **-13.69%** | - | **-2.96%** |

[a] $\gamma = 0.0001$. [b] $\gamma = 0.002$. [c] $\gamma = 0.02$. [d] $\gamma = 0.05$.

*Table S3. Evaluation of layer-to-layer knowledge mixing method (LKM) across different knowledge distillation strengths ($\gamma$) using MXMNet. Comparison uses the QM9 dataset. The average $\Delta E_{mae}$ reduces and reaches a minimum at 0.02 and then rises off after that.*

| Task | Unit | Mean absolute error (Lower is better) | | | | | | | | |
|---|---|---|---|---|---|---|---|---|---|---|
| | | Knowledge distillation strength ($\gamma$) | | | | | | | | |
| | | 0.0 (Baseline) | 0.001 | 0.002 | 0.005 | 0.01 | 0.02 | 0.05 | 1 | 100 |
| Mu | D | 0.0384 | 0.0369 | 0.0361 | 0.0352 | 0.0338 | 0.0325 | 0.0322 | **0.0311** | 0.0592 |
| Alpha | $a_0^3$ | 0.0524 | 0.0496 | 0.0493 | 0.0482 | 0.0479 | 0.0479 | **0.0475** | 0.0544 | 0.0609 |
| HOMO | meV | 23.8 | 22.9 | 22.6 | 22.9 | 22.6 | **22.1** | 22.8 | 24.5 | 33.0 |
| LUMO | meV | 18.9 | 19.2 | 18.9 | 18.7 | 18.2 | **17.9** | 18.3 | 19.9 | 23.9 |
| Gap | meV | 40.1 | 40.3 | 39.7 | 39.6 | 38.2 | 38.1 | **37.9** | 40.5 | 53.6 |
| $<R^2>$ | $a_0^2$ | 0.530 | 0.469 | 0.448 | 0.482 | 0.470 | 0.481 | 0.444 | **0.414** | 0.469 |
| ZPVE | meV | **1.24** | 1.29 | 1.26 | 1.26 | 1.28 | 1.30 | 1.31 | 1.31 | 1.33 |
| $U_0$ | meV | 8.99 | 7.51 | **7.22** | 7.53 | 7.35 | 7.54 | 7.72 | 8.26 | 8.79 |
| U | meV | 8.53 | 7.77 | 7.62 | **7.32** | 7.44 | 7.75 | 8.68 | 8.85 | 9.17 |
| H | meV | 7.78 | 7.73 | **7.21** | 7.51 | 7.61 | 7.39 | 7.33 | 8.65 | 8.88 |
| G | meV | 9.73 | 8.20 | 8.11 | 7.95 | 7.93 | **7.92** | 8.02 | 9.15 | 9.27 |
| $C_v$ | $\frac{cal}{mol\ K}$ | **0.024** | **0.024** | **0.024** | **0.024** | **0.024** | **0.024** | **0.024** | **0.024** | 0.026 |
| Average $\Delta E_{mae}$ (%) | | - | -5.01% | -7.18% | -6.84% | -7.82% | **-7.89%** | -7.05% | -1.78% | +15.66% |

*Table S4. Evaluation of layer-to-layer knowledge mixing method (LKM) across different number of GNN layers using MXMNet at a fixed knowledge distillation strength ($\gamma = 0.05$). Comparison uses the QM9 dataset. The LKM method gives accuracy improvement for all number of layers. More intriguingly, the average $\Delta E_{mae}$ decreases with the number of layers, demonstrating the effectiveness of knowledge mixing.*

|  |  | Mean absolute error (lower is better) | | | | | |
| --- | --- | --- | --- | --- | --- | --- | --- |
|  |  | 2 layers | | 4 layers | | 6 layers | |
| Task | Unit | Base | Base+LKM (ours) | Base | Base+LKM (ours) | Base | Base+LKM (ours) |
| Mu | D | 0.0462 | **0.0401** | 0.0412 | **0.0336** | 0.0384 | **0.0322** |
| Alpha | $a_0^3$ | 0.0559 | **0.0501** | 0.0527 | **0.0475** | 0.0525 | **0.0475** |
| HOMO | meV | **26.5** | 26.6 | 24.7 | **23.4** | 23.8 | **22.8** |
| LUMO | meV | 22.2 | **21.0** | 20.7 | **19.2** | 18.9 | **18.3** |
| Gap | meV | 45.9 | **42.8** | 42.2 | **38.9** | 40.1 | **37.9** |
| <$R^2$> | $a_0^2$ | 0.603 | **0.536** | 0.495 | **0.461** | 0.530 | **0.444** |
| ZPVE | meV | 1.28 | **1.25** | **1.25** | 1.29 | **1.24** | 1.31 |
| $U_0$ | meV | 8.21 | **7.82** | 7.75 | **7.38** | 8.99 | **7.72** |
| U | meV | 8.08 | **7.99** | 7.79 | **7.25** | **8.53** | 8.68 |
| H | meV | 8.19 | **7.67** | 7.85 | **7.59** | 7.78 | **7.33** |
| G | meV | 8.53 | **8.31** | 8.72 | **7.86** | 9.73 | **8.02** |
| $C_v$ | $\frac{cal}{mol\ K}$ | 0.0262 | **0.0249** | 0.0249 | **0.0242** | 0.0237 | **0.0236** |
| Number of winning tasks |  | 1/12 | 11/12 | 1/12 | 11/12 | 2/12 | 10/12 |
| average $\Delta E_{mae}$ (%) |  | - | -5.71% | - | -6.67% | - | -7.11% |

**For Table of Contents Only**

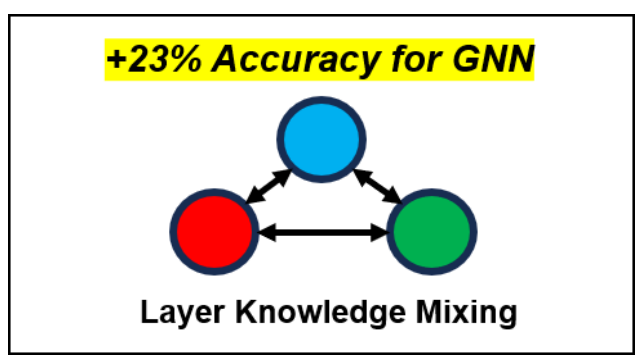